\documentclass[10pt,twocolumn,letterpaper]{article}
\usepackage{iccv}
\usepackage{times}
\usepackage{algorithm}
\usepackage{algpseudocode}
\usepackage{amsmath}
\usepackage{amssymb}
\usepackage{graphicx}
\usepackage{todonotes}
\usepackage[pagebackref=true,breaklinks=true,letterpaper=true,colorlinks,bookmarks=false]{hyperref}
\usepackage{cleveref}
\usepackage{setspace}
\usepackage{array}
\usepackage[numbers,sort]{natbib} 

\presetkeys{todonotes}{inline,backgroundcolor=yellow}{}

\makeatletter
\renewcommand{\paragraph}{%
  \@startsection{paragraph}{4}%
  {\z@}{0em}{-1em}%
  {\normalfont\normalsize\bfseries}
}
\makeatother

\iccvfinalcopy
\def\iccvPaperID{281}
\ificcvfinal\fi

\title{Learning to Discover Novel Visual Categories via Deep Transfer Clustering}
\author{Kai Han\hspace{2em} Andrea Vedaldi \hspace{2em} Andrew Zisserman\\
Visual Geometry Group, University of Oxford\\
{\tt\small \{khan, vedaldi, az\}@robots.ox.ac.uk}}

\begin{document}
\maketitle
\begin{abstract}
We consider the problem of discovering novel object categories in an image collection.
While these images are unlabelled, we also assume prior knowledge of related but different image classes.
We use such prior knowledge to reduce the ambiguity of clustering, and improve the quality of the newly discovered classes.
Our contributions are twofold.
The first contribution is to extend Deep Embedded Clustering to a transfer learning setting; we also improve the algorithm by introducing a representation bottleneck, temporal ensembling, and consistency.
The second contribution is a method to estimate the number of classes in the unlabelled data.
This also transfers knowledge from the known classes, using them as probes to diagnose different choices for the number of classes in the unlabelled subset.
We thoroughly evaluate our method, substantially outperforming state-of-the-art techniques in a large number of benchmarks, including ImageNet, OmniGlot, CIFAR-100, CIFAR-10, and SVHN.
\end{abstract}
\section{Introduction}\label{s:intro}

With modern supervised learning methods, machines can recognize thousands of visual categories with high reliability; in fact, machines can \emph{outperform} individual humans when performance depends on extensive domain-specific knowledge as required for example to recognize hundreds of species of dogs in ImageNet~\cite{deng09imagnet}.
However, it is also clear that machines are still far behind human intelligence in some fundamental ways.
A prime example is the fact that good recognition performance can only be obtained if computer vision algorithms are \emph{manually supervised}.
Modern machine learning methods have little to offer in an open world setting, in which image categories are \emph{not} defined a-priori, or for which no labelled data is available.
In other words, machines lack an ability to structure data automatically, understanding concepts such as object categories without external supervision.

\begin{figure}[t]
\centering
\includegraphics[width=0.95\linewidth]{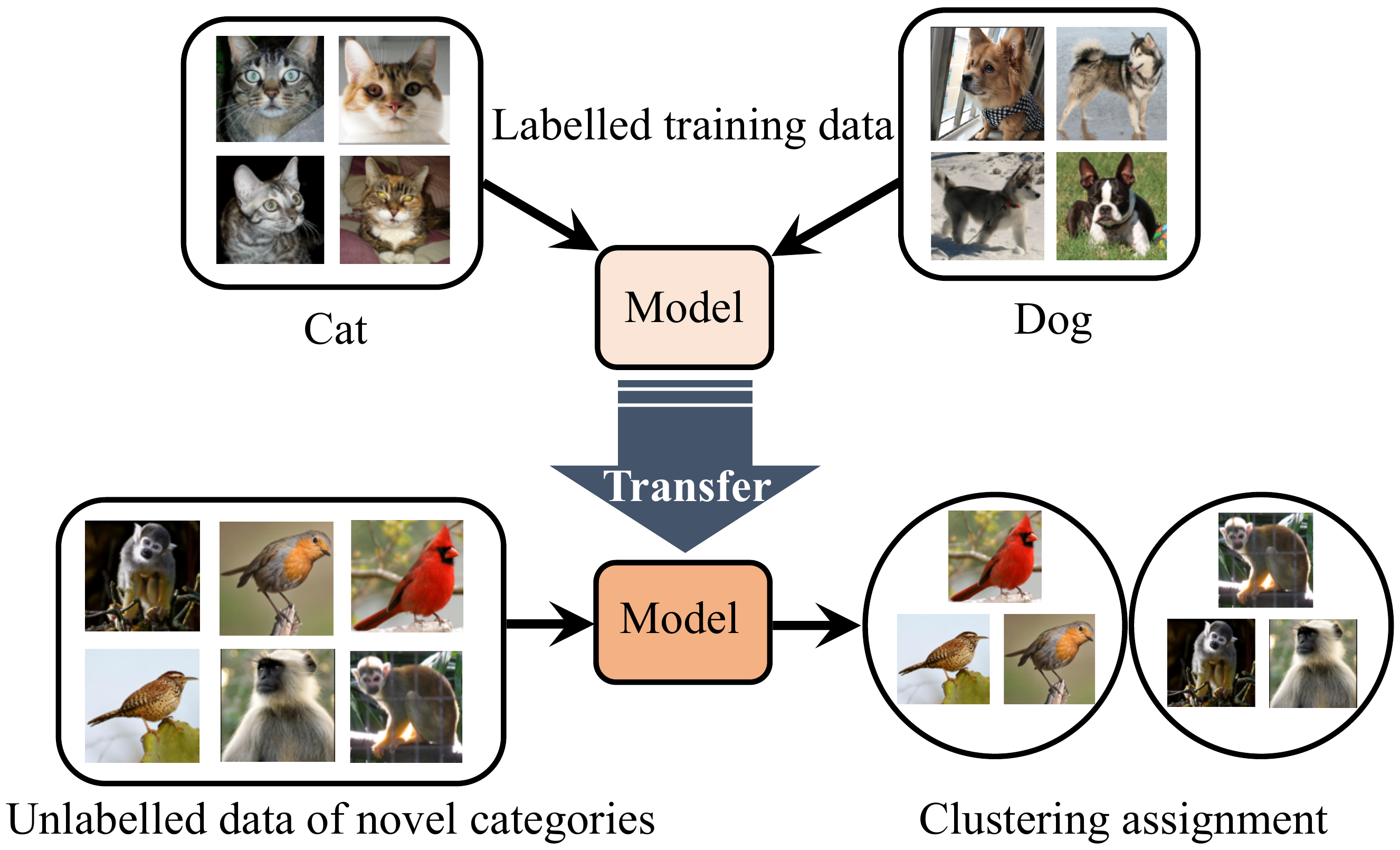}
\caption{Learning to discover novel visual categories via deep transfer clustering. We first train a model with labelled images (e.g., cat and dog). The model is then applied to images of unlabelled novel categories (e.g., bird and monkey), which transfers the knowledge learned from the labelled images to the unlabelled images. With such transferred knowledge, our model can then simultaneously learn a feature representation and the clustering assignment for the unlabelled images of novel categories.}\label{f:splash}
\end{figure}

In this paper, we study the problem of \emph{discovering and recognizing} visual categories automatically.
However, rather than considering a fully unsupervised setting, we assume that the machine already possesses certain knowledge about some of the categories in the world.
Then, given additional images that belong to \emph{new} categories, the problem is to tell how many new categories there are and to learn to recognize them.
The aim is to guide this process by transferring knowledge from the old classes to the new ones (see~\cref{f:splash}).

This approach is motivated by the following observation.
Unlike existing machine learning models, a child can easily tell an unseen animal category (e.g.,
bird) after learning a few other (seen) animal categories (e.g., cat,
dog); and an adult wandering around a zoo or wildlife park can
effortlessly discover new categories of animals (e.g., okapi) based on
the many categories previously learnt.
In fact, while we can manually annotate \emph{some} categories in the world, we cannot annotate them
all, even in relatively restricted settings.  For example, consider
the problem of recognizing products in supermarkets for the purpose of
market research:  hundreds of new products
are introduced every week and providing manual annotations for all is
hopelessly expensive.  However, an algorithm can draw on knowledge
of several thousand products in order to discover new
ones as soon as they enter the data stream.

This problem lies at the intersection of three widely-studied areas:  semi-supervised learning~\cite{chapelle2006semi}, transfer learning~\cite{Pan10transfer,weiss2016asurvey}, and clustering~\cite{Aggarwal13cluster}.
However, it has not been extensively addressed in any of them.
In semi-supervised learning, labelled and unlabelled data contain the same categories, an assumption that is not valid in our case.
Moreover, semi-supervised learning has been shown to perform poorly if the unlabelled data is contaminated with new categories~\cite{oliver2018realistic}, which is problematic in our case.
In transfer learning~\cite{Pan10transfer}, a model may be trained on one set of categories and then fine-tuned to recognize different categories, but both source and target datasets are annotated, whereas in our case the target dataset is unlabelled.
Our problem is more similar to clustering~\cite{Aggarwal13cluster}, extensively studied in machine learning, since the goal is to discover classes without supervision.
However, our goal is also to leverage knowledge of other classes to improve the discovery of the new ones.
Since classes are a high-level abstraction, discovering them automatically is  challenging, and perhaps impossible since there are many criteria that could be used to cluster data (e.g.,~we may equally well cluster objects by color, size, or shape).
Knowledge about some classes is not only a realistic assumption, but also indispensable to narrow down the meaning of clustering.

Our contribution is a method that can discover and learn new object categories in unlabelled data while leveraging knowledge of related categories.
This method has two components.
The first is a modification of a recent deep clustering approach, Deep Embedded Clustering (DEC)~\cite{Xie16_DEC}, that can cluster data while learning a data representation at the same time.
The purpose of the modification is to allow clustering to be guided by the known classes.
We also extend the algorithm by introducing a representational bottleneck, temporal ensembling, and consistency, which boost its performance considerably.

However, this method still requires to know the number of new categories in the unlabelled data, which is not a realistic assumption in many applications.
So, the second component is a mechanism to estimate the number of classes.
This also transfers knowledge from the set of known classes.
The idea is to use part of the known classes as a probe set, adding them to the unlabelled set pretending that part of them are unlabelled, and then running the clustering algorithm described above on the extended unlabelled dataset.
This allows to cross-validate the number of classes to pick, according to the clustering accuracy on the probe set as well as a cluster quality index on the unlabelled set, resulting in a reliable estimate of the true number of unlabelled classes.

We empirically demonstrate the strength of our approach, utilizing public benchmarks such as ImageNet, OmniGlot, CIFAR-100, CIFAR-10, and SVHN, and outperforming competitors by a substantial margin in all cases.
Our code can be found at \url{http://www.robots.ox.ac.uk/~vgg/research/DTC}.

\section{Related work}
Our work is related to semi-supervised learning, transfer learning, and clustering. These three areas are widely studied, and it is out the scope of this paper to review all of them. We briefly review the most representative and related works below in each area.

Semi-supervised learning (SSL)~\cite{chapelle2006semi,oliver2018realistic,Sylvestre2019semi} aims to solve a closed-set classification problem in which part of the data is labelled while the rest is not. In the context of SSL, both the labelled data and unlabelled data share the same categories, while this assumption does not hold in our case. A comprehensive study of recent SSL methods can be found in~\cite{oliver2018realistic}. The consistency based SSL methods (e.g., \cite{laine2016temporal, tarvainen2017mean}) have been shown to achieve promising results. 
Laine and Aila~\cite{laine2016temporal} proposed to incorporate the unlabelled data during training by the consistency between the predictions of a data sample and its transformed counterpart, which they call the $\Pi$ model, or by the consistency between current prediction and the temporal ensembling prediction. Instead of keeping a temporal ensembling prediction, Tarvainen and Valpola~\cite{tarvainen2017mean} proposed to maintain a temporal ensembling model, and enforces the consistency between predictions of the main model and the temporal ensembling model. 

In transfer learning~\cite{Pan10transfer,weiss2016asurvey,tan2018asurvey}, a model is first trained on one labelled dataset, and then fine-tuned with another labelled dataset,  containing different categories. Our case is similar to transfer learning in the sense that we also transfer knowledge from a source dataset to a target dataset, though our target dataset is unlabelled. With the advent of deep learning, the most common way of transfer learning nowadays is to fine-tune models pretrained on ImageNet~\cite{deng09imagnet} for specific tasks with labelled data. However, in our case, no labels are available for the new task.

Clustering~\cite{Aggarwal13cluster} has long been studied in machine learning. A number of classic works (e.g., $k$-means~\cite{MackQueen67_Kmeans}, mean-shift~\cite{Comaniciu02meanshift}) have been widely applied in many applications. Recently, there have been more and more works on clustering in the deep learning literature (e.g.,~\cite{Xie16_DEC,Chang_2017_ICCV,Dizaji2017deepclustering,Yang17towards,yang2016joint}). Among them, Deep Embedded Clustering (DEC)~\cite{Xie16_DEC} appears to be one of most promising learning based clustering approaches. It can simultaneously cluster the data and learn a proper data representation. It is trained in two phases. The first phase trains an autoencoder using reconstruction loss, and the second phase finetunes the encoder of the autoencoder with an auxiliary target distribution. However, it does not take the available labelled data of seen categories into account, thus the performance is still far from satisfactory. 

Our work is also related to metric learning~\cite{song2017deepmetric,song2016deepmetric,Sohn2016improved} and domain adaptation~\cite{wang18deepvisual}. 
Actually, we \emph{build} on metric learning, as the latter is used for initialization. However, most metric learning methods are unable to exploit unlabelled data, while our work can automatically adjust the embeddings space on unlabelled data.
More importantly, our task requires producing a partition of the data (a discrete decision), whereas metric learning only produces a continuous data embedding, and
converting the latter to discrete classes is often not trivial.
Domain adaptation aims to resolve the domain discrepancy between source and target datasets (e.g., digital SLR camera images vs web camera images), while generally assuming a shared class space. Thus, the source and target data are on different manifolds. In our case, the unlabelled data belongs to novel categories without any labels, and the unlabelled data are on the same manifold with the labelled data, which is a more practical but more challenging scenario.

To our knowledge, the most related works to ours are~\cite{Hsu18_L2C} and~\cite{Hsu19_MCL}, in terms of considering novel visual category discovery as a deep transfer clustering task. In~\cite{Hsu18_L2C}, Hsu \emph{et al.} introduced a Constrained Clustering Network (CCN) which is trained in two stages. In the first stage, a binary classification model is trained on labelled data to measure pair-wise similarity of images. In the second stage, a clustering model is trained on unlabelled data by using the output of the binary classification model as supervision. 
The network is trained with a Kullback-Leibler divergence based contrastive loss (KCL). In~\cite{Hsu19_MCL}, the CCN is improved by replacing KCL with a new loss called Meta Classification Likelihood (MCL). In addition, Huang \emph{et al.}~\cite{Huang2019centroid} recently introduced Centroid Networks for few-shot clustering, which cluster $K\times M$ unlabeled images into $K$ clusters with $M$ images each after training on labeled data. 

\section{Deep transfer clustering}\label{s:method}

We propose a method for data clustering:
given as input an unlabelled dataset $D^u=\{x_i^u,i=1,\dots,M\}$, usually of images, the goal is to produce as output class assignments $y_i^u \in \{1,\dots, K\}$, where the number of different classes $K$ is unknown.
Since there can be multiple equally-valid criteria for clustering data, making a choice depends on the application.
Thus, we also assume we have a labelled dataset $D^l=\{(x_i^l,y_i^l),i=1,\dots,N\}$ where class assignments $y_i^l\in\{1,\dots,L\}$ are known.

The classes in this labelled set differ, in identity and number, from the classes in the unlabelled set.
Hence the goal is to learn from the labelled data not its specific classes, but what properties make a good class in general, so that this knowledge can be used to discover new classes and their number in the unlabelled data.

We propose a method with two components.
The first is an extension of a deep clustering algorithm that can transfer knowledge from a known set of classes to a new one (\cref{s:clustering});
the second is a method to reliably estimate the number of unlabelled classes $K$ (\cref{s:k}).

\subsection{Transfer clustering and representation learning}\label{s:clustering}

At its core, our method is based on a deep clustering algorithm that clusters the data while simultaneously learning a good data representation.
We extract this representation by applying a neural network $f_\theta$ to the data, obtaining embedding vectors $z = f_\theta(x) \in \mathbb{R}^d$.
The representation is initialised using the labelled data, and then fine-tuned using the unlabelled data.
This is done via \emph{deep embedded clustering} (DEC) of~\cite{Xie16_DEC} with three important modifications: the method is extended to account for labelled data, to include a tight bottleneck to improve generalization, and to incorporate temporal ensembling and consistency, which also contribute to its stability and performance.
An overview of our approach is given in \cref{alg:known_k}.

\subsubsection{Joint clustering and representation learning}\label{s:dec}

In this section we summarise DEC~\cite{Xie16_DEC} as this algorithm lies at the core of our approach.
In DEC, similar to $k$-means, clusters are represented by a collection of vectors or prototypes $U = \{ \mu_k,k=1,\dots,K \}$ representing the cluster ``centers''.
However, differently from $k$-means, the goal is not only to determine the clusters, but also to learn the data representation $f_\theta$.

Naively combining representation learning, which is a discriminative task, and clustering, which is a generative one, is challenging.
For instance, directly minimizing the $k$-means objective function would immediately collapse the learned representation vectors to the closest cluster centers.
DEC~\cite{Xie16_DEC} addresses this problem by slowly annealing cluster centers and data representation.

In order to do so, let $p(k|i)$ be the probability of assigning data point $i\in\{1,\dots,N\}$ to cluster $k\in\{1,\dots,K\}$.
DEC uses the following parameteterization of this conditional distribution by assuming a Student's $t$ distribution:


\begin{equation}\label{eq:q_ij}
p(k|i)
\propto
\left(1
+
\frac{\|z_{i}-\mu_{k}\|^{2}}{\alpha}
\right)^{-\frac{\alpha+1}{2}}.
\end{equation}
Further assuming that data indices are sampled uniformly (i.e.~$p(i)=1/N$), we can write the joint distribution $p(i,k) = p(k|i)/N$.




In order to anneal to a good solution, instead of maximizing the likelihood of the model $p$ directly, we match the model to a suitably-shaped distribution $q$.
This is done by minimizing the KL divergence between joint distributions $q(i,k)= q(k|i)/N$ and $p(i,k) = p(k|i)/N$, given by
\begin{equation}\label{e:kl}
E(q)
=
KL(q||p)
=
\frac{1}{N}\sum_{i=1}^N\sum_{k=1}^K	q(k|i) \log \frac{q(k|i)}{p(k|i)}.
\end{equation}

It remains to show how to construct the target distribution $q$ as a progressively sharper version of the current distribution $p$.
Concretely, this is done by setting
$$
q(k|i)
\propto
p(k|i) \cdot p(i|k).
$$
In this manner the assignment of image $i$ to cluster $k$ is reinforced when the current distribution $p$ assigns a high probability of going from $i$ to $k$ \emph{as well as} of going from $k$ to $i$.
The latter has an equalization effect as the probability of sampling data point $i$ in cluster $k$ is high only if the cluster is not too large. Using Bayes rule for $p(i|k)$, the   expression can be rewritten as
\begin{equation}\label{e:target}
q(k|i)
\propto
\frac{p(k|i)^2}{\sum_{i=1}^N p(k|i)}.
\end{equation}
Hence the target distribution is constructed by first raising $p(k|i)$ to the second power, which sharpens it, and then normalizing by the frequency per cluster, which balances it.

In practice, \cref{e:kl} is minimized in alternate-optimization fashion.
Namely, fixing a target distribution $q(k|i)$, the representation $f_\theta$ is optimized using stochastic gradient descent or a similar method to minimize~\cref{e:kl} for a certain number of iteration, usually corresponding to a complete sweep over the available training data  (an epoch).
\Cref{e:target} is then used to sharpen the target distribution and the process is repeated.


\begin{algorithm}[t]
\caption{Transfer clustering with known cardinality}\label{alg:known_k}
\begin{algorithmic}[1]


\State \textbf{Initialization:}\\ 
Train the feature extractor $f_{\theta}$ on the labelled data $D^l$.
Apply $f_\theta$ to the unlabelled data $D_u$ to extract features, use PCA to reduce the latter to $K$ dimensions, and use $K$-means to initialize the centers $U$. Incorporate the PCA as a final linear layer in $f_\theta$. Construct target distributions $q$. 
\State \textbf{Warm-up training:}
\For{$t \in \{1, \dots, N_\text{warm-up}\}$} 
  \State Train $\theta$ and $U$ on $D_u$ using $q$ as target.
\EndFor
\State Update target distributions $q$.
\State \textbf{Main loop:}
\For{$t \in \{1, \dots, N_\text{train}\}$} 
  \State Train $\theta$ and $U$ on $D_u$ using $q$ as target.
  \State Update target distributions $q$.
\EndFor
\State Predict $p(k|i)$ for $i=1, \dots, M$ and $k=1, \dots, K$.
\State Return $y^u_i = \operatornamewithlimits{argmax}_{k} p(k|i)$ for $i=1, \dots, M$.
\end{algorithmic} 
\end{algorithm}

\paragraph{Transferring knowledge from known categories.}\label{s:labelled}
The clustering algorithm described above is entirely unsupervised.
However, our goal is to aid the discovery of new classes by leveraging a certain number of known classes.
We capture such information in the image representation $f_\theta$, which is pre-trained on the labelled dataset $\mathcal{D}^l$ using a metric learning approach.
In order to train $f_\theta$, one can use the cross-entropy loss, the triplet loss or the prototypical loss, depending on what is the best supervised approach for the specific data.

\paragraph{Bottleneck.}\label{s:bottleneck}
\Cref{alg:known_k} requires an initial setting for the cluster centers $U$.
We obtain this initialization by running the $k$-means algorithm on the set of features $\mathcal{Z}^u=\{z_i = f_\theta(x^u_i), i=1,\dots,M\}$ extracted from the unlabelled data.
However, we found this step to perform much better by introducing a step of dimensionality reduction in the feature representation $z_i\in\mathbb{R}^d$.
To this end, PCA is applied to the feature vectors $\mathcal{Z}^u$, resulting in a dimensionality reduction layer $\hat z_i = A z_i + b$. Importantly, we retain a number of components equal to the number of unlabelled classes $K$, so that $A \in\mathbb{R}^{K\times d}$.
This linear layer is then added permanently as the head of the deep network, and the parameters $A, b$ are further fine-tuned during clustering together with the other parameters.

\subsubsection{Temporal ensembling and consistency}\label{sec:te}
The key idea of DEC is to slowly anneal clusters to learn a meaningful partition of the data. Here, we propose a modification of DEC that can further improve the smoothness of the annealing process via temporal ensembling~\cite{laine2016temporal}.
To apply temporal ensembling to DEC, the clustering models $p$ computed at different epochs are aggregated by maintaining an exponential moving average (EMA) of the previous distributions.

In more detail, we first accumulate the network predictions $p$ into an ensemble prediction $P$ via
\begin{equation}\label{eq:qti}
P^{t}(k|i)= \beta \cdot P^{t-1}(k|i)+(1-\beta)\cdot p^{t}(k|i),
\end{equation}
where $\beta$ is a momentum term controlling how far the ensemble reaches into training history, and $t$ indicates the time step.
To correct the zero initialization of the EMA~\cite{laine2016temporal},  $P^t$ is rescaled to obtain the smoothed model distribution
\begin{equation}\label{eq:qti_cor}
\tilde{p}^t(k|i) = \frac{1}{1-\beta^t} \cdot P^{t}(k|i).
\end{equation}
\Cref{eq:qti_cor} is plugged into~\cref{e:target} to obtain a new target distribution $\tilde q^t(k|i)$.
In turn, this defines a variant of~\cref{e:kl} that is then optimized to learn the model.

Consistency constraints have been shown to be effective in SSL (e.g., \cite{laine2016temporal,tarvainen2017mean}). A consistency constraint can be incorporated by enforcing the predictions of a data sample and its transformed counterpart (which can be obtained by applying data transformation such as random cropping and horizontal flipping on the original data sample)
to be close (known as the $\Pi$ model in SSL), or by enforcing the prediction of a data sample and its temporal ensemble prediction to be close. Such consistency constraints can also be used to improve our method. After introducing consistency, the loss in~\cref{e:kl} now becomes
\begin{equation}
\begin{aligned}
E(q)=
& \frac{1}{N}\sum_{i=1}^N\sum_{k=1}^K	q(k|i) \log \frac{q(k|i)}{p(k|i)} \\
& + \omega(t) \frac{1}{NK}\sum_{i=1}^N\sum_{k=1}^K \| p(k|i) - p'(k|i)\|^2,
\end{aligned}
\end{equation}
where $p'(k|i)$ is either the prediction of the transformed sample or the temporal ensemble prediction $\tilde{p}^t(k|i)$, and $\omega(t)$ is a ramp-up function as used in~\cite{laine2016temporal,tarvainen2017mean} to gradually increase the weight of the consistency constraint from 0 to 1.

\subsection{Estimating the number of classes}\label{s:k}

\begin{algorithm}[t]
\caption{Estimating the number of classes}\label{alg:unknown_k}
\begin{algorithmic}[1]
\State \textbf{Preparation:}
\State Split the probe set $D^l_r$ into $D^l_{ra}$ and $D^l_{rv}$.
\State Extract features of $D^l_r$ and $D^u$ using $f_\theta$.
\State \textbf{Main loop:}
\For{$0\leq K \leq K_\text{max}$}
  \State
  Run $k$-means on $D^l_r \cup D^u$ assuming $L_r+K$ classes in semi-supervised mode (i.e.~forcing data in $D^l_{ra}$ to map to the ground-truth class labels).
  \State Compute ACC for $D^l_{rv}$ and CVI for $D^u$.
\EndFor
\State \textbf{Obtain optimal:}
\State Let $K^*_a$ be the value of $K$ that maximise ACC for $D^l_{rv}$ and $K^*_v$ be the value that maximise CVI for $D^u$ and let $\hat K = (K^*_a + K^*_v)/2$. Run semi-supervised $K$-means on $D^l_r \cup D^u$ again assuming $L_r+\hat K$ classes.
\State \textbf{Remove outliers:}
\State Look at the resulting clusters in $D^u$ and drop any that has a mass less than $\tau$ of the largest cluster. Output the number of remaining clusters.
\end{algorithmic} 
\end{algorithm}

So far, we have assumed that the number of classes $K$ in the unlabelled data is known, but this is usually not the case in real applications.
Here we propose a new approach to estimate the number of classes in the unlabelled data by making use of labelled probe classes.
The probe classes are combined with the unlabelled data and the resulting set is clustered using $k$-means multiple times, varying the number of classes.
The resulting clusters are then examined by computing two quality indices, one of which checks how well the probe classes, for which ground truth data is available, have been identified.
The number of categories is then estimated to be the one that maximizes these quality indices.

In more details, we first split the $L$ known classes into a probe subset $D^l_r$ of $L_r$ classes and a training subset $D^l \setminus D^l_r$ containing the remaining $L-L_r$ classes. The $L-L_r$ classes are used for supervised feature representation learning, while the $L_r$ probe classes are combined with the unlabelled data for class number estimation. We then further split the $L_r$ probe classes into a subset $D^l_{ra}$ of $L^a_r$ classes and a subset $D^l_{rv}$ of $L^v_r$ classes (e.g.,  $L^a_r : L^v_r = 4:1$), which we call anchor probe set and validation probe set respectively.
We then run a constrained (semi-supervised) $k$-means on $D^l_r \cup D^u$ to estimate the number of classes in $D^u$. Namely, during $k$-means, we force images in the anchor probe set $D^l_{ra}$ to map to clusters following their ground-truth labels, while images in the validation probe set $D^l_{rv}$  are considered as additional ``unlabelled'' data.
We launch this constrained $k$-means multiple times by sweeping the number of total categories $C$ in $D^l_r \cup D^u$, and measure the constrained clustering quality on  $D^l_r \cup D^u$.
We consider two quality indices, given below, for each value of $C$.
The first measures the cluster quality in the $L^v_r$ labelled validation probe set, whereas the second measures the quality in the unlabelled data $\mathcal{D}^u$.
Each index is used to determine an optimal number of classes and the results are averaged.
Finally, $k$-means is run one last time with this value as number of classes and any outlier cluster in $D^u$, defined as containing less than $\tau$ (e.g., $\tau = 1\%$) the mass of the largest clusters, are dropped.
The details are given in~\cref{alg:unknown_k}.

\noindent \textbf{Cluster quality indices}.
We measure our clustering for class number estimation with two indices.
The first index is the \emph{average clustering accuracy} (ACC), which is applicable to the $L^v_r$ labelled classes in the validation probe set $D^l_{rv}$ and is given by
\begin{equation}\label{e:acc}
  \operatornamewithlimits{max}_{g \in \operatorname{Sym}(L^v_r)}
  \frac{1}{N} \sum_{i=1}^{N} \mathbf{1}\left\{\bar{y}_{i}=g\left(y_{i}\right)\right\},
\end{equation}
where $\overline{y}_{i}$ and ${y}_{i}$ denote the ground-truth label and clustering assignment for each data point $x_i\in D^l_{rv}$ and $\operatorname{Sym}(L^v_r)$ is the group of permutations of $L^v_r$ elements (as a clustering algorithm recovers clusters in an arbitrary order).

The other index is a \emph{cluster validity index} (CVI)~\cite{Arbelaitz12cluster} which, by capturing notions such as  intra-cluster cohesion vs inter-cluster separation, is applicable to the unlabelled data $D^u$.
There are several CVI metrics, such as Silhouette~\cite{Rousseeuw87Silhouettes}, Dunn~\cite{Dunn74dunn}, Davies–Bouldin~\cite{Davies79pami}, and Calinski-Harabasz~\cite{Cali74cluster}; while no metric is uniformly the best, the Silhouette index generally works well~\cite{Arbelaitz12cluster,Bezdek98cluster}, and we found it to be a good choice for our case too.
This index is given by
\begin{equation}\label{e:silhouette}
\sum_{x\in\mathcal{D}^u} \frac{b(x) - a(x)}{\max\{a(x),b(x)\}},
\end{equation}
where $x$ is a data sample, $a(x)$ is the average distance between $x$ and all other data samples within the same cluster, and $b(x)$ is the smallest average distance of $x$ to all points in any other cluster (of which $x$ is not a member).

\section{Experimental results}

We assess two scenarios over multiple benchmarks: first, where the number of new classes is known  for OmniGlot, ImageNet, CIFAR-10, CIFAR-100 and SVHN; and second, where the number of new classes is unknown for OmniGlot, ImageNet and CIFAR-100. For the unknown scenario we separate a probe set from the labelled classes.


\subsection{Data and experimental details}\label{s:exp_detail}

\noindent \textbf{OmniGlot}~\cite{Lake15omnniglot}.
This dataset contains 1,623 handwritten characters from 50 different alphabets. It is divided into a 30-alphabet (964 characters) subset called \textit{background} set and a 20-alphabet (659 characters) subset called \textit{evaluation} set.
Each character is considered as one category and has 20 example images. We use the \textit{background} and \textit{evaluation} sets as labelled and unlabelled data, respectively.
To experiment with an unknown number of classes, we randomly hold out 5 alphabets from the \textit{background} set (169 characters in total) to use as probes for~\cref{alg:unknown_k}, leaving the remaining 795 characters to learn the feature extractor.

\noindent \textbf{ImageNet}~\cite{deng09imagnet}.
ImageNet contains 1,000 classes and about 1,000 example images per class.
We follow~\cite{vinyals2015matching} and split the data into two subsets containing 882 and 118 classes respectively.
Following~\cite{Hsu18_L2C,Hsu19_MCL}, we consider the 882-class subset as labelled data, and use three randomly sampled 30-class subsets from the remaining 118-class subset as unlabelled data.
To experiment with an unknown number of classes, we randomly hold out 82 classes from the 882-class subset as probes, leaving the remaining 800 classes for training the feature extractor. 

\noindent \textbf{CIFAR-10/CIFAR-100}~\cite{Krizhevsky09cifar}.
CIFAR-10 contains 50,000 training images and 10,000 test images from 10 classes. Each image has a size of $32\times 32$.  We split the training images into labelled and unlabelled subsets.
In particular, we consider the images of the first 5 categories (i.e., airplane, automobile, bird, cat, deer) as the labelled set, while the remaining 5 categories (i.e., dog, frog, horse, ship, truck) as the unlabelled set.
CIFAR-100 is similar to CIFAR-10, except it has 10 times less images per class.
We consider the first 80 classes as labelled data, and the last 10 classes as unlabelled data, leaving 10 classes as probe set for category number estimation on unlabelled data.

\noindent \textbf{SVHN}~\cite{Netzer2011svhn}.
SVHN contains 73,257 images of digits for training and 26,032 images for testing.
We split the 73,257 training digits into labelled and unlabelled subsets.
Namely, we consider the images of digits 0-4 as the labelled set, and the images of 5-9 as the unlabelled set.
The labelled set contains 45,349 images, while the unlabelled set contains 27,908 images.

\noindent \textbf{Evaluation metrics.}
We adopt the conventionally used clustering accuracy (ACC) and normalized mutual information (NMI)~\cite{Strehl02cluster} to evaluate the clustering performance of our approach.
Both metrics are valued in the range of $[0, 1]$ and higher values mean better performance.
We measure error in the estimation of the number of novel categories as $|K_{gt} - K_{est}|$, where $K_{gt}$ and $K_{est}$ denote the ground-truth and estimated number of categories, respectively.

\noindent \textbf{Network architectures.} 
For a fair comparison, we follow~\cite{Hsu18_L2C,Hsu19_MCL} and use a 6-layer VGG like architecture~\cite{Simonyan15vgg} for OmniGlot and CIFAR-100, and a ResNet18~\cite{he2016deep} for ImageNet and all other datasets.

\noindent \textbf{Training configurations.}
OmniGlot is widely used in the context of few-shot learning due to the very large number of categories it contains and the small number of example images per category.
Hence, in order to train the feature extractor on the \textit{background} set of OmniGlot we use the prototypical loss~\cite{snell2017prototypical}, one of the best methods for few-shot learning.
We train the feature extractor with a batch size of 200, forming batches by randomly sampling 20 categories and including 10 images per category.
For each category, 5 images are used as supporting data to calculate the prototypes while the remaining 5 images are used as query samples.
We use Adam optimizer with a learning rate of 0.001 for 200 epochs.
We then finetune $f_\theta$ and train the bottleneck and the cluster centers $U$ for each alphabet in the \textit{evaluation} set.
For warm-up (in \cref{alg:known_k}), the Adam optimizer is used with a learning rate of 0.001, and trained for 10 epochs without updating the target distribution. Afterwards, training continues for another 90 epochs updating the target distribution per epoch.
For ImageNet and other datasets, which are widely used in supervised image classification tasks, we pre-train the feature extractor using the cross-entropy loss on the labelled subsets. Following common practice, we then remove the last layer of the classification network and use the the rest of the model as our feature extractor. 

In our experiment on ImageNet, we take the pretrained ImageNet$_{882}$ classification network of~\cite{Hsu19_MCL} as our initial feature extractor.
For the case when the number of novel categories is unknown, we train a ImageNet$_{800}$ classification network as our initial feature extractor.
We use SGD with an initial learning rate of 0.1, which is divided by 10 every 30 epochs, for 90 epochs.
For warm-up, the feature extractor, together with the bottleneck and cluster centers, are trained for 10 epochs by SGD with a learning rate of 0.1; then, we train for further 60 epochs updating the target distribution per epoch.
Experiments on other datasets follow a similar configuration.
Our results on all datasets are averaged over 10 runs, except ImageNet, which is averaged over 3 runs using different unlabelled subsets following~\cite{Hsu18_L2C,Hsu19_MCL}.

\subsection{Learning with a known number of categories}
\label{s:result_known_k}
\begin{table*}[ht]
\footnotesize
\caption{Visual category discovery (known number of categories).}\label{tab:self_compare}
\centering
\begin{tabular}{l|cc|cc|cc|cc|cc}
  \hline
    & \multicolumn{2}{c}{CIFAR-10} 
    & \multicolumn{2}{c}{CIFAR-100} 
    & \multicolumn{2}{c}{SVHN}
    & \multicolumn{2}{c}{OmniGlot} 
    & \multicolumn{2}{c}{ImageNet} \\
  \hline
  Method & ACC & NMI & ACC & NMI & ACC & NMI & ACC & NMI & ACC & NMI \\
  \hline
  $k$-means~\cite{MackQueen67_Kmeans}  & 65.5\% & 0.422 & 66.2\% & 0.555 & 42.6\% & 0.182 & 77.2\% & 0.888  & 71.9\%  & 0.713 \\
  DTC-Baseline & 74.9\% & 0.572 & 72.1\% & 0.630 & 57.6\% & 0.348 & 87.9\% & 0.933 & \textbf{78.3}\% & 0.790 \\
  DTC-$\Pi$ & \textbf{87.5}\% & \textbf{0.735} & 70.6\% & 0.605 & \textbf{60.9}\% & \textbf{0.419} & \textbf{89.0}\% & \textbf{0.949} & 76.7\% & 0.767 \\
  DTC-TE & 82.8\% & 0.661 & \textbf{72.8}\% & \textbf{0.634} & 55.8\% & 0.353 & 87.8\% & 0.931 & 78.2\% & \textbf{0.791} \\
  DTC-TEP & 75.2\% & 0.591 & 72.5\% & 0.632 & 55.4\% & 0.329 & 87.8\% & 0.932 & \textbf{78.3}\% & \textbf{0.791} \\
  \hline
\end{tabular}
\end{table*}

In~\cref{tab:self_compare} we compare variants of our Deep Transfer Clustering (DTC) approach with the temporal ensembling and consistency constraints as introduced in~\cref{sec:te}, namely, DTC-Baseline (our model trained using DEC loss),  DTC-$\Pi$ (our model trained using DEC loss with consistency constraint between predictions of a sample and its transformed counterpart), DTC-TE (our model trained using DEC loss with consistency constraint between current prediction and temporal ensemble prediction of each sample), and DTC-TEP (our mode trained using DEC loss with targets constructed from temporal ensemble predictions). We only apply standard data augmentation of random crop and horizontal flip in our experiment. To measure the performance of metric learning based initialization, we also show the results of $k$-means~\cite{MackQueen67_Kmeans} on the features of unlabelled data produced by our feature extractor trained with the labelled data. $k$-means shows reasonably good results on clustering the unlabelled data using the model trained on labelled data, indicating that the model can transfer useful information to cluster data of unlabelled novel categories. All variants of our approach substantially outperform $k$-means, showing that our approach can effectively finetune the feature extractor and cluster the data. 
DTC-$\Pi$ appears to be the most effective one for CIFAR-10, SVHN, and OmniGlot. The consistency constraints makes a huge improvement for CIFAR-10 (e.g., ACC $74.9\%\xrightarrow{}87.5\%$) and SVHN (e.g., ACC $57.6\%\xrightarrow{}60.9\%$). When it comes to the more challenging datasets, CIFAR-100 and ImageNet, DTC-TE and DTC-TEP appear to be the most effective with ACC of $72.8\%$ and $78.3\%$ respectively.

\begin{figure}[htb]
\centering
\input{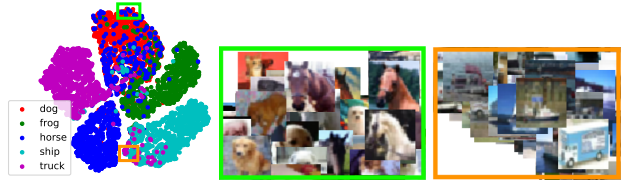}
\caption{Representation visualization on CIFAR-10. Left: t-SNE projection on our learned features of unlabelled data (colored with GT labels); Middle: failure cases of clustering horses as dogs; Right: failure cases of clustering trucks as ships.}\label{f:quali_failure}
\end{figure}

We visualize the t-SNE projection of our learened feature on unlabelled subset of CIFAR-10 in~\cref{f:quali_failure}. It can be seen that our learned representation is sufficiently discriminative for different novel classes, clearly demonstrating that our approach can effectively discover novel categories. We also show some failure cases where there exist some confusion between dogs and horse heads (due to a similar pose and color) in the green selection, and between trucks and ships in the orange selection (the trucks are either parked next to the sea, or have a similar color with the sea).

\begin{table}
\centering
\footnotesize
\caption{Results on OmniGlot and ImageNet with known number of categories.}\label{tab:omniglot_known}
\begin{tabular}{l|cc|cc}
  \hline
    & \multicolumn{2}{c}{OmniGlot} 
    & \multicolumn{2}{c}{ImageNet} \\
\hline
Method & ACC  &NMI  & ACC  &NMI  \\
\hline
$k$-means~\cite{MackQueen67_Kmeans} & 21.7\%  & 0.353 & 71.9\%  & 0.713\\
LPNMF~\cite{Cai09_LPNMF} & 22.2\%  & 0.372 & 43.0\%  & 0.526 \\
LSC~\cite{Chen11_LSC} & 23.6\% & 0.376 & 73.3\% & 0.733\\
\hline
KCL~\cite{Hsu18_L2C} & 82.4\%  & 0.889 & 73.8\%  & 0.750\\
MCL~\cite{Hsu19_MCL} & 83.3\%  & 0.897 & 74.4\%  & 0.762\\
Centroid Networks~\cite{Huang2019centroid} & 86.6\% & - & - & -\\
\hline
DTC & \textbf{89.0}\% & \textbf{0.949} & \textbf{78.3}\% & \textbf{0.791}\\
\hline
\end{tabular}
\end{table}


We compare our approach with traditional methods as well as state-of-the-art learning based methods on OmniGlot and ImageNet in~\cref{tab:omniglot_known}. We use the same 6-layer VGG like architecture as KCL~\cite{Hsu18_L2C}, MCL~\cite{Hsu19_MCL} and Centroid Networks~\cite{Huang2019centroid} for comparison on OmniGlot, and use the same ResNet18 as KCL and MCL for comparison on ImageNet. The results of traditional methods are those reported in~\cite{Hsu19_MCL} using raw images for OmniGlot and pretrained features for ImageNet.
All these methods are applied by assuming the number of categories to be known.
It is worth noting that Centroid Networks~\cite{Huang2019centroid} also assumes the clusters to be of uniform size.
This assumption, although not practical in real application, is beneficial when experimenting with OmniGlot, since each category contains exactly 20 images.
For both datasets, our method outperforms existing methods in both ACC (89.0\% vs 86.6\%) and NMI (0.949 vs 0.897). Unlike KCL and MCL, our method does not need to maintain an extra model to provide a pseudo-supervision signal for the clustering model.

In addition, we also compare with KCL and MCL on CIFAR-10, CIFAR-100, and SVHN in~\cref{tab:kcl_mcl} based on their officially-released code. Our method consistently outperforms KCL and MCL on these datasets, which further verifies the effectiveness of our approach.

\begin{table}[ht]
\centering
\footnotesize
\caption{Comparison with KCL and MCL on CIFAR-10/CIFAR-100/SVHN.}\label{tab:kcl_mcl}
\begin{tabular}{l|cc|cc|cc}
  \hline
    & \multicolumn{2}{c}{CIFAR-10}
    & \multicolumn{2}{c}{CIFAR-100}
    & \multicolumn{2}{c}{SVHN} \\
  \hline
   & ACC & NMI & ACC & NMI & ACC & NMI \\
  \hline
  KCL~\cite{Hsu18_L2C} & 66.5\% & 0.438 & 27.4\% & 0.151 & 21.4\% & 0.001 \\
  MCL~\cite{Hsu19_MCL} & 64.2\% & 0.398 & 32.7\% & 0.202 & 38.6\% & 0.138 \\
  DTC & \textbf{87.5}\% & \textbf{0.735} & \textbf{72.8}\% & \textbf{0.634} & \textbf{60.9}\% & \textbf{0.419} \\
  \hline
\end{tabular}
\end{table}

\subsection{Finding the number of novel categories}
\label{s:results_unknown_k}

We now experiment under the more challenging (and realistic) scenario where the number of categories in unlabelled data is unknown.
KCL amd MCL assume the number of categories to be a large value (i.e., 100) instead of estimating the number of categories explicitly.
By contrast, we choose to estimate the number of categories before running the transfer clustering algorithm using~\cref{alg:unknown_k} (with $K_\text{max} = 100$ for all our experiments) and only then apply~\cref{alg:known_k} to find the clusters.
Results for novel category number estimation are reported in~\cref{tab:imagenet_k_est}. The average error is less than 5 for all of three datasets, which validates the effectiveness of our approach.
In~\cref{tab:omniglot_unknown}, we show the clustering results on OmniGlot and ImageNet for~\cref{alg:known_k}, with these estimates for the number of novel categories, and also compare with other methods. 
The results of traditional methods are those reported in~\cite{Hsu19_MCL} using raw images for OmniGlot and pretrained features for ImageNet. 
In both datasets, our approach achieves the best results, outperforming previous state-of-the-art by 6.8\% and  6.1\% ACC on OmniGlot and ImageNet respectively.

We also experiment on KCL and MCL by using our estimated number of clusters on OmniGlot and ImageNet (see~\cref{tab:kcl_mcl_w_our_k}).
With this augmentation, both KCL amd MCL improve significantly in term of ACC, and are similar in term of NMI, indicating that our category number estimation method can also be beneficial for other methods. Our method still significantly outperforms the augmented KCL and MCL on all metrics.


\begin{table}
\centering
\footnotesize
\caption{Category number estimation results.}\label{tab:imagenet_k_est}
\begin{tabular}{l|ccc}
\hline
Data & GT  & Ours & Error \\
\hline
OmniGlot & 20-47 & 22-51 & 4.60 \\
ImageNet$_\text{A, B, C}$ & \{30, 30, 30\} & \{34, 31, 32\} & 2.33\\
CIFAR-100 & 10 & 11 & 1\\
\hline
\end{tabular}
\end{table}

\begin{table}
\footnotesize
\centering
\caption{Results on OmniGlot and ImageNet with unknown number of categories.}\label{tab:omniglot_unknown}
\begin{tabular}{l|cc|cc}
  \hline
    & \multicolumn{2}{c}{OmniGlot} 
    & \multicolumn{2}{c}{ImageNet} \\
\hline
Method & ACC  &NMI  & ACC  &NMI  \\
\hline
$k$-means~\cite{MackQueen67_Kmeans} & 18.9\% & 0.464 & 34.5\%  & 0.671\\
LPNMF~\cite{Cai09_LPNMF}  & 16.3\%  & 0.498 & 21.8\%  & 0.500 \\
LSC~\cite{Chen11_LSC}  & 18.0\%  & 0.500 & 33.5\% & 0.655\\
\hline
KCL~\cite{Hsu18_L2C} & 78.1\% & 0.874 & 65.2\%  & 0.715\\
MCL~\cite{Hsu19_MCL} & 80.2\% & 0.893 & 71.5\%  & 0.765\\
\hline
DTC &  \textbf{87.0}\% &  \textbf{0.945} & \textbf{77.6}\% & \textbf{0.786} \\
\hline
\end{tabular}
\end{table}


\begin{table}[h]
\centering
\footnotesize
\caption{KCL and MCL with our category number estimation.}\label{tab:kcl_mcl_w_our_k}
\begin{tabular}{l|cc|cc}
  \hline
    & \multicolumn{2}{c}{OmniGlot}
    & \multicolumn{2}{c}{ImageNet} \\
  \hline
   & ACC & NMI & ACC & NMI \\
  \hline
  KCL~\cite{Hsu18_L2C} & 78.1\% & 0.874 & 65.2\% & 0.715\\
  KCL~\cite{Hsu18_L2C} w/ our $k$ & 80.3\% & 0.875 & 71.4\% & 0.740\\
  MCL~\cite{Hsu19_MCL} & 80.2\% & 0.893 & 71.5\% & 0.765\\
  MCL~\cite{Hsu19_MCL} w/ our $k$ & 80.5\% & 0.879 & 72.9\% & 0.752\\
  DTC & \textbf{87.0}\% & \textbf{0.945} & \textbf{77.6}\% & \textbf{0.786}\\
  \hline
\end{tabular}
\end{table}

\subsection{Transfer from ImageNet pretrained model}
\label{s:results_transfer_imagenet}
The most common way of transfer learning with modern deep convolutional neural networks is to use ImageNet pretrained models. Here, we explore the potential of leveraging the ImageNet pretrained model to transfer features for novel category discovery. In particular, we take the ImageNet pretrained model as our feature extractor, and adopt our transfer clustering model on a new dataset. We experiment with CIFAR-10 and the results are shown in \cref{tab:imagenet_transfer}. Instead of considering only part of the categories as unlabelled data, we consider the whole CIFAR-10 training set as unlabelled data here. Similar as before, our deep transfering clustering model equipped with temporal ensembling or consistency constraints consistently outperform $k$-means and our baseline model. DTC-$\Pi$ performs the best in term of ACC and DTC-TE performs the best in term of NMI. We also experimented with SVHN, however we do not have much success on it. This is likely due to the small correlation between ImageNet and SVHN. This result is consistent with that of semi-supervised learning (SSL)~\cite{oliver2018realistic}.
Using an ImageNet pretrained model, SSL can achieve  reasonably good  performance on CIFAR-10, but not on SVHN, which shows that the correlation between source data and target data is important for SSL. 
Our results corroborate that, to successfully transfer knowledge from the pretrained models for deep transfer clustering, the labelled data and unlabelled data should be closely related.

\begin{table}
\centering
\footnotesize
\caption{Results of transferring from ImageNet to CIFAR-10. }\label{tab:imagenet_transfer}
\begin{tabular}{l|ccc}
\hline
& ACC & NMI  \\
\hline
$k$-means~\cite{MackQueen67_Kmeans} & 71.0\%  & 0.639 \\
DTC-Baseline & 76.9\%  & 0.729 \\
DTC-$\Pi$ & \textbf{78.9}\%  & 0.753 \\
DTC-TE & 78.5\%  & \textbf{0.755} \\
DTC-TEP & 77.4\% & 0.734 \\
\hline
\end{tabular}
\end{table}
\section{Conclusion}
We have introduced a simple and effective approach for novel visual category discovery in unlabelled data, by considering it as a deep transfer clustering problem.
Our method can simultaneously learn a data representation and cluster the unlabelled data of novel visual categories, while leveraging knowledge of related categories in labelled data.
We have also proposed a novel method to reliably estimate the number of categories in unlabelled data by transferring cluster prior knowledge using labelled probe data.
We have thoroughly evaluated our method on public benchmarks, and it substantially outperformed state-of-the-art techniques in both known and unknown category number cases, demonstrating the effectiveness of our approach.
\paragraph{Acknowledgments.}
\begin{spacing}{1.0}
We are grateful to EPSRC Programme Grant Seebibyte EP/M013774/1 and ERC StG IDIU-638009 for support.
\end{spacing}

{\small\bibliographystyle{ieee_fullname}\bibliography{novel}}

\onecolumn
\appendix
\section*{Appendices}

\section{Bottleneck dimension}

\begin{figure}[htb]
\centering
\includegraphics[width=0.55\linewidth]{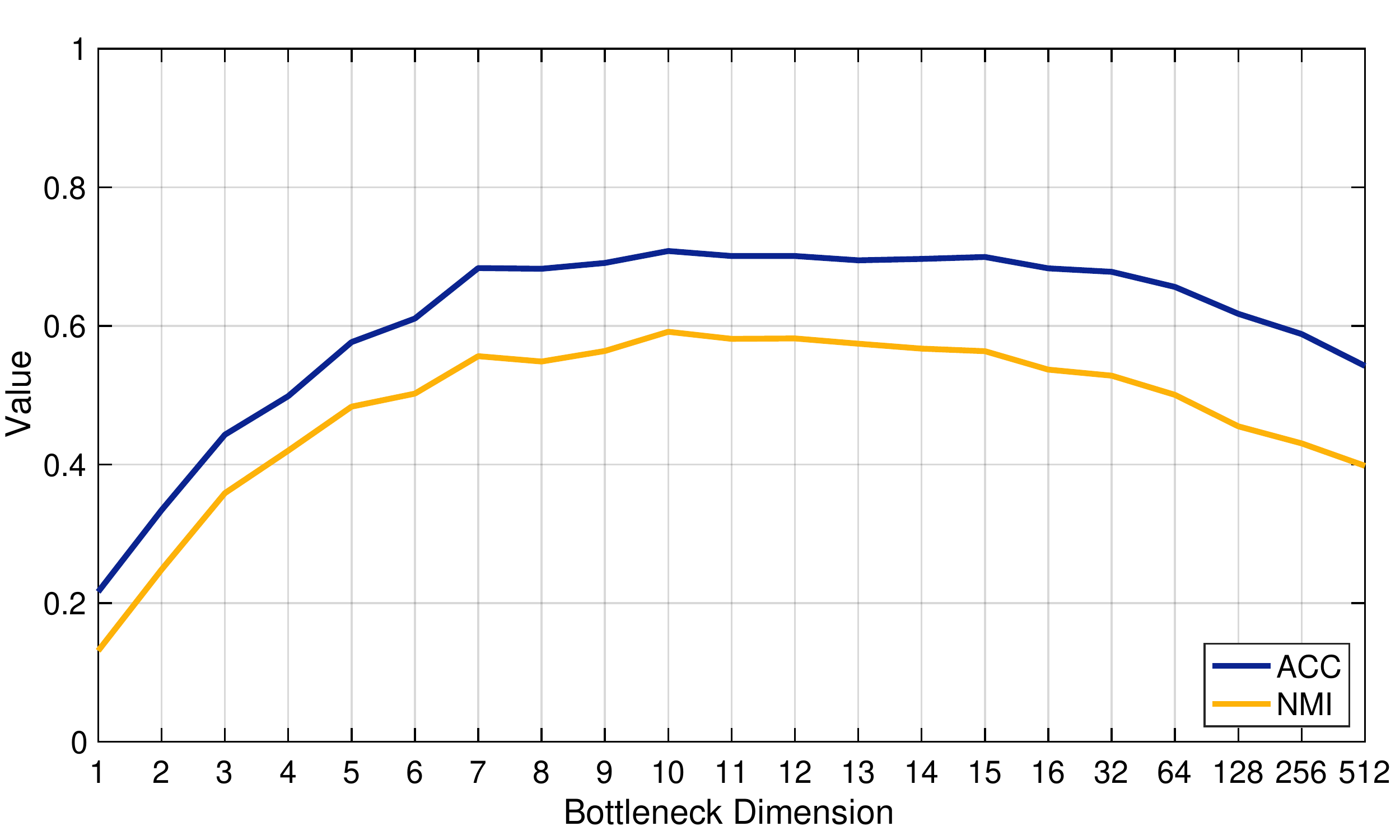}
\caption{ACC and NMI w.r.t. different bottleneck dimensions.}\label{f:feat_d}
\end{figure}

As described in section 3.1.1 of our paper, we introduce a bottleneck layer $\{A, b\}$ to reduce to dimension of the learned representation from $d$ (e.g., $d=512$ for ResNet18 which is used in the experiment) to $c$, where $A \in \mathbb{R}^{c \times d}$ and $b \in \mathbb{R}^{c \times 1}$. To verify different choices of $c$ for $A \in \mathbb{R}^{c \times d}$ in the bottleneck, we experiment with the 10-class unlabelled subset of CIFAR100 by varying $c$. In particular, we train Ours-Baseline model with different $c$ in the bottleneck. The ACC and NMI are shown in~\cref{f:feat_d}. It can be seen that our model is not very sensitive to the choice of $c$, especially when $c$ is slightly larger than the number of unlabelled categories  $K$ ($K=10$ in this experiment). We find that setting $c=K$ is a good choice for the bottleneck, since it gives the best ACC and NMI in our experiment.


\section{Number of clusters}

\begin{figure}[htb]
\centering
\includegraphics[width=0.55\linewidth]{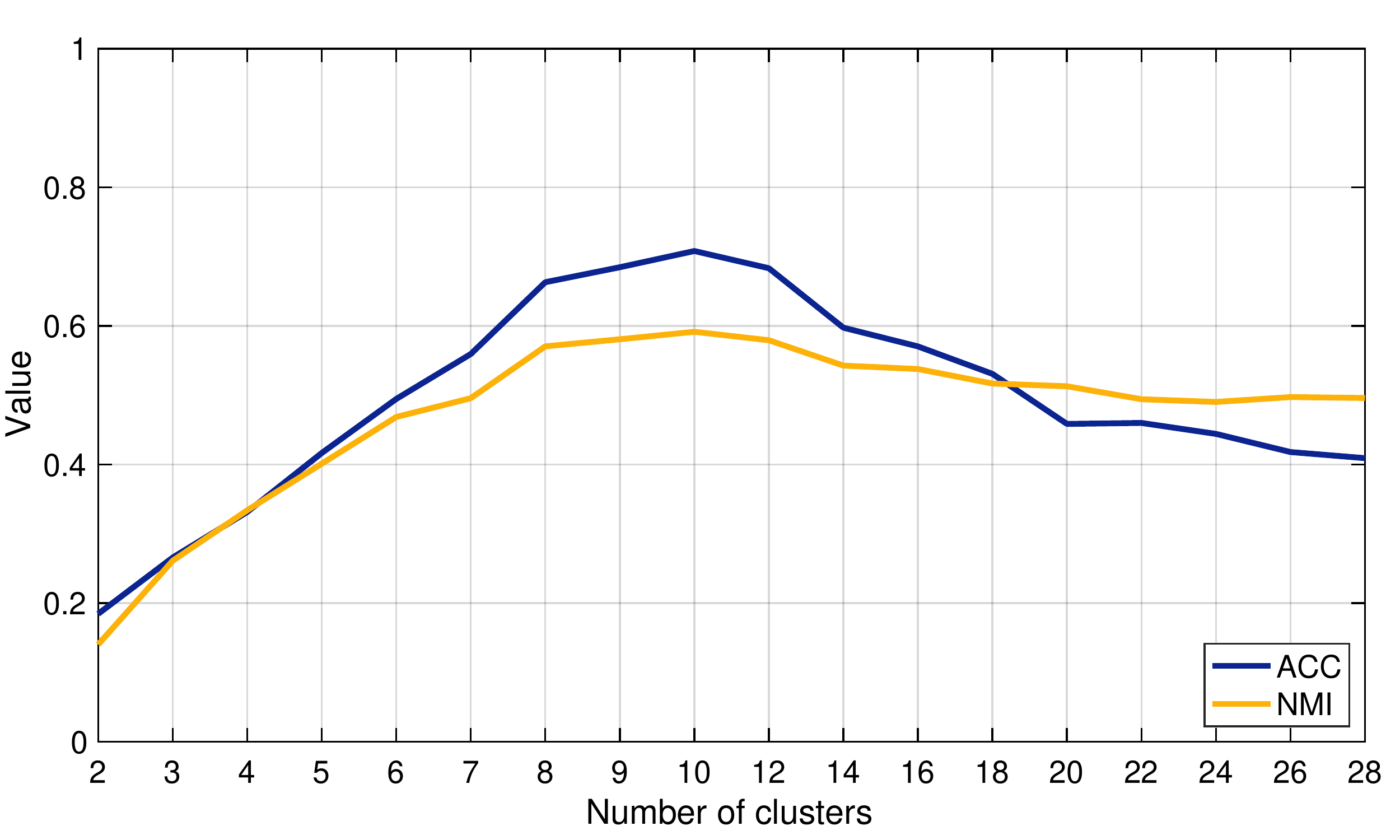}
\caption{ACC and NMI w.r.t. different number of clusters.}\label{f:different_k}
\end{figure}

Our transfer clustering model requires the number of novel categories to be known. However, this is not always the case in real applications. When it is unknown, we can use our algorithm introduced in section 3.2 to estimate it. For the 10-class unlabelled subset of CIFAR100, our algorithm gives an estimate of 12 which is very close to the ground truth (i.e., 10). We show the results of setting different number of clusters for our transfer clustering model in~\cref{f:different_k}. It can be seen that both ACC and NMI decrease if the estimated number of categories is different from the ground truth. While a larger number is more preferable than a smaller number, since ACC and NMI decrease faster with smaller numbers than lager numbers.

\clearpage

\section{Representation visualization}
\begin{figure*}[h]
   \centering
   \tabcolsep=0.02cm
   
   	   \renewcommand{\arraystretch}{0.25}
   \begin{tabular}{
         >{\centering\arraybackslash} m{0.33\textwidth}
         >{\centering\arraybackslash} m{0.33\textwidth}
         >{\centering\arraybackslash} m{0.33\textwidth}}
    \includegraphics[width=1\linewidth]{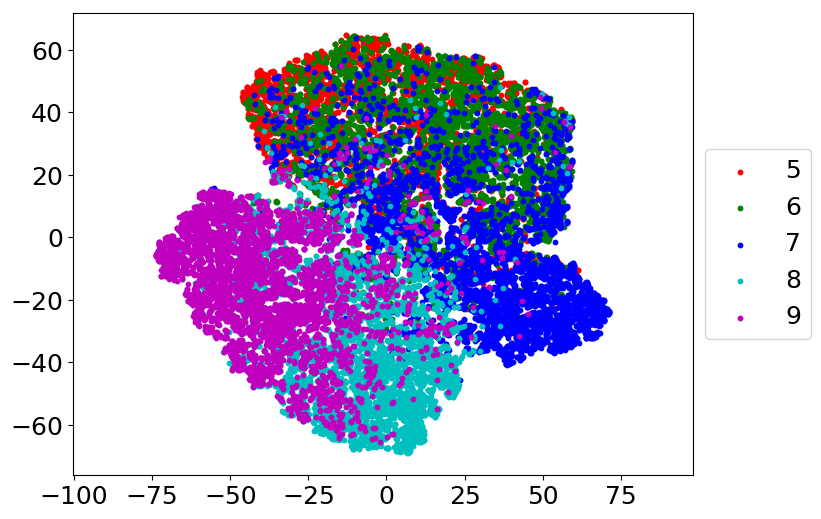} &
    \includegraphics[width=1\linewidth]{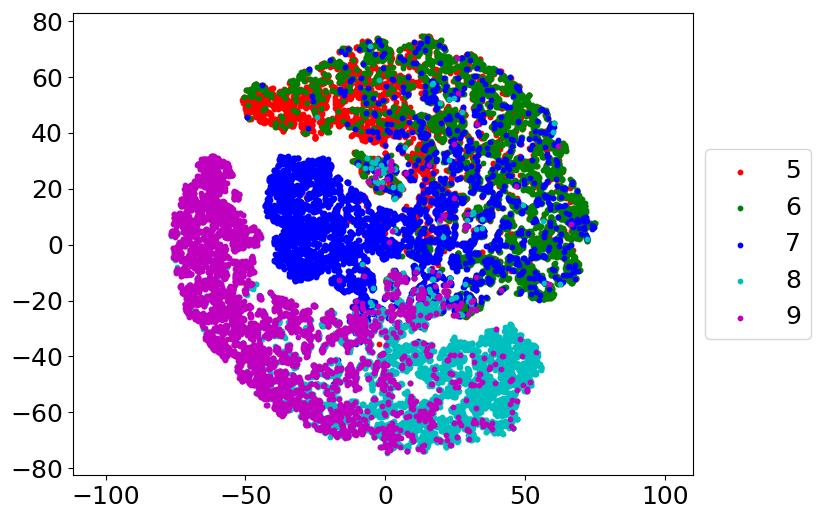}&
    \includegraphics[width=1\linewidth]{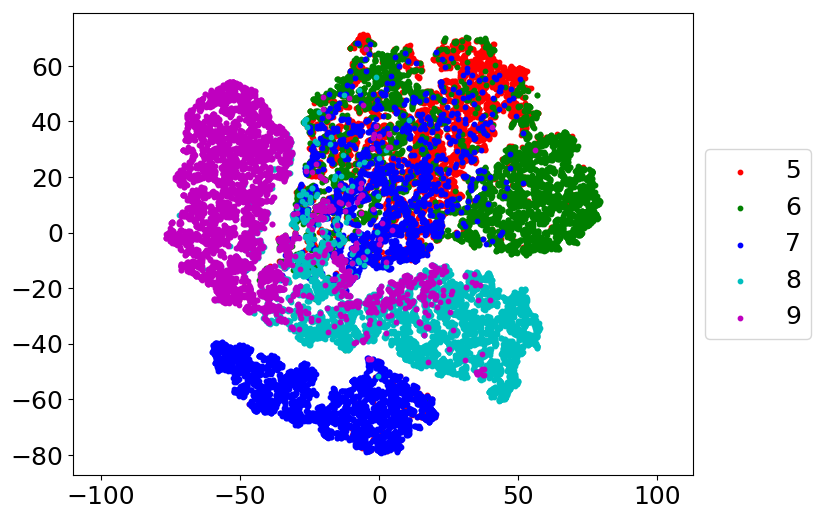} \\
    (a) init & (b) epoch 0 & (c) epoch 10\\
    \includegraphics[width=1\linewidth]{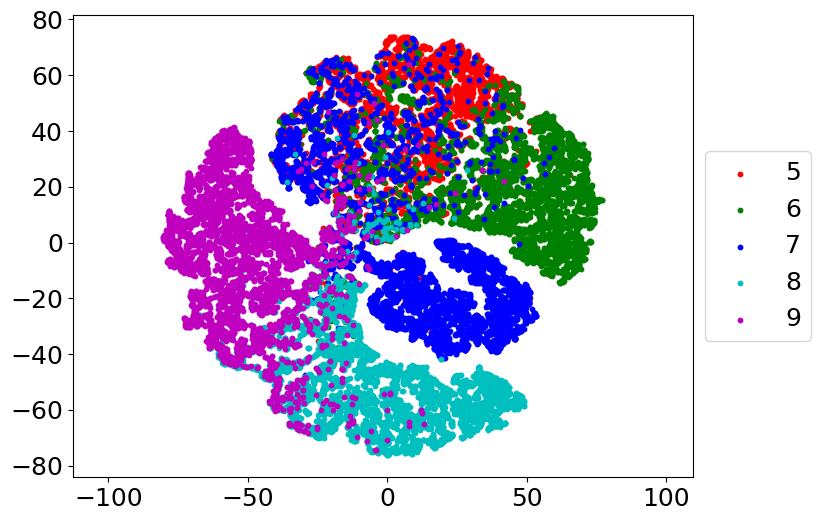} &
    \includegraphics[width=1\linewidth]{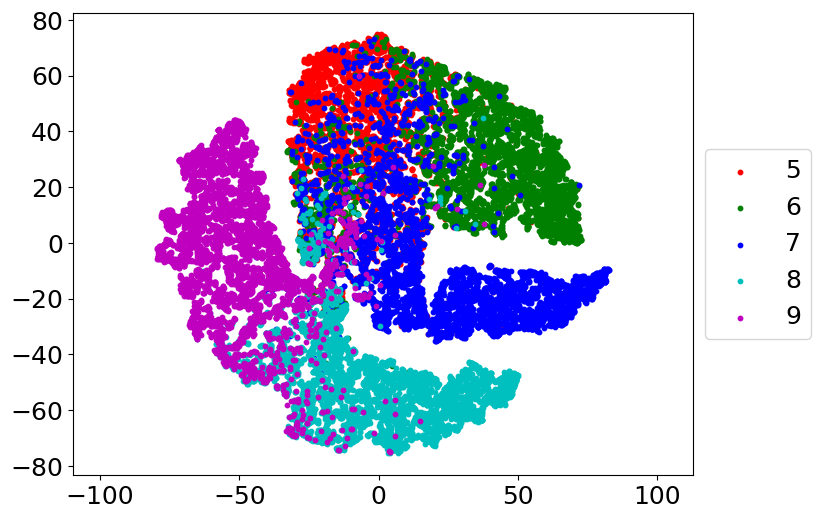} &
    \includegraphics[width=1\linewidth]{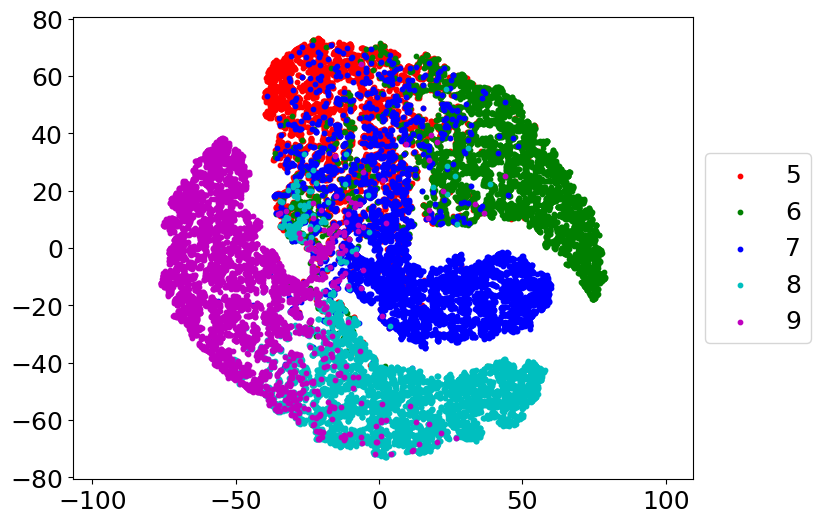} \\
    (d) epoch 20 & (e) epoch 30 & (f) epoch 40 \\
    \includegraphics[width=1\linewidth]{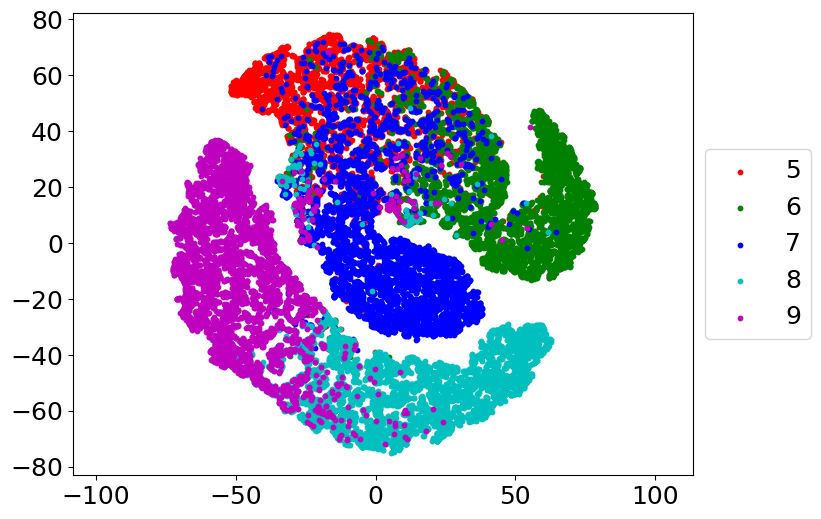} &
    \includegraphics[width=1\linewidth]{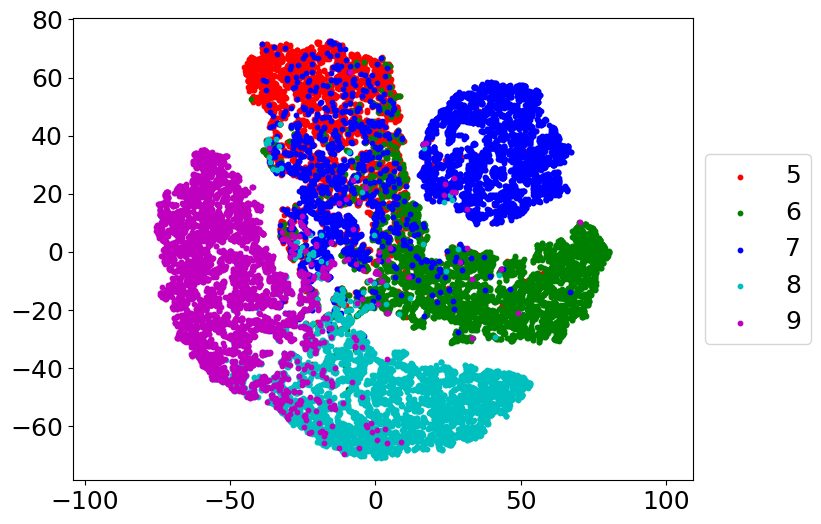} &
    \includegraphics[width=1\linewidth]{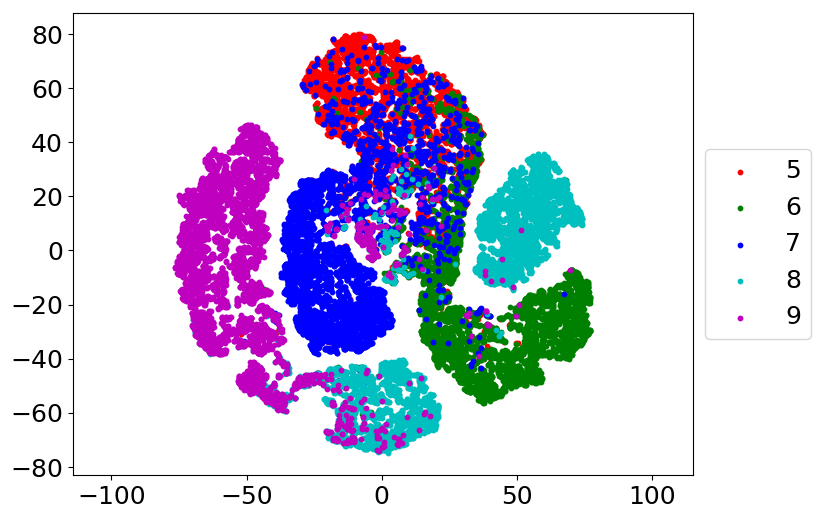} \\
    (g) epoch 50 & (h) epoch 60 & (i) epoch 70\\
    \includegraphics[width=1\linewidth]{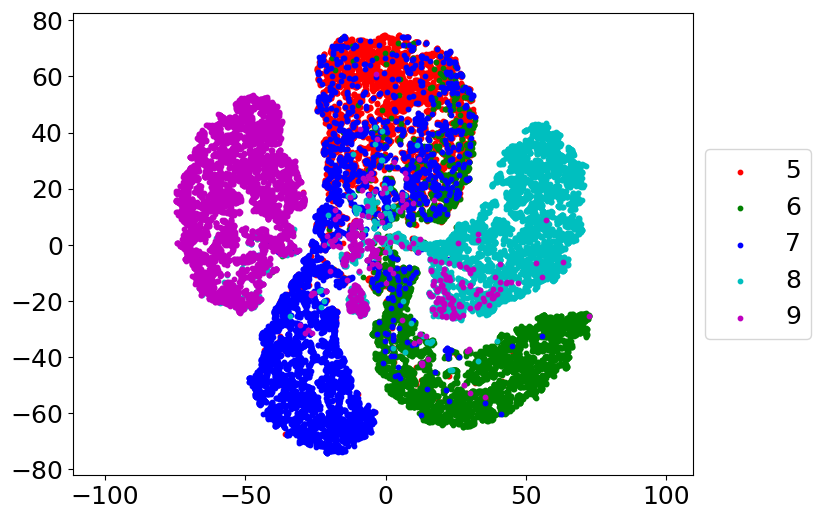} &
    \includegraphics[width=1\linewidth]{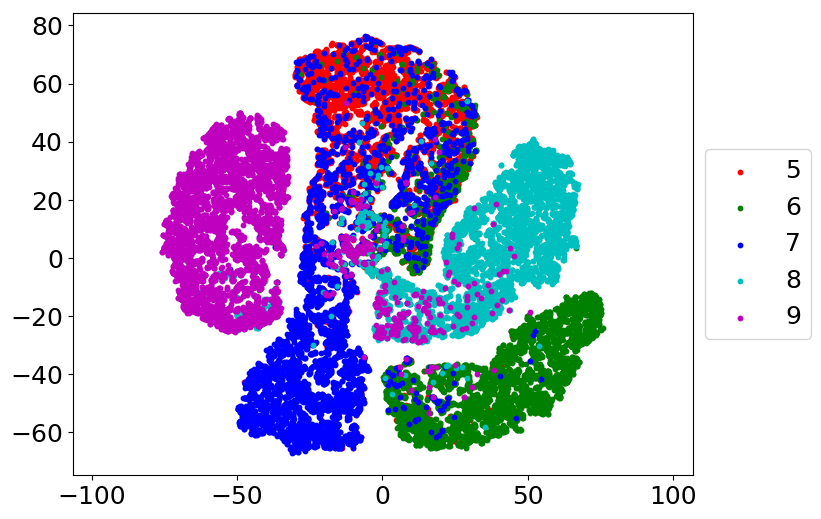} &
    \includegraphics[width=1\linewidth]{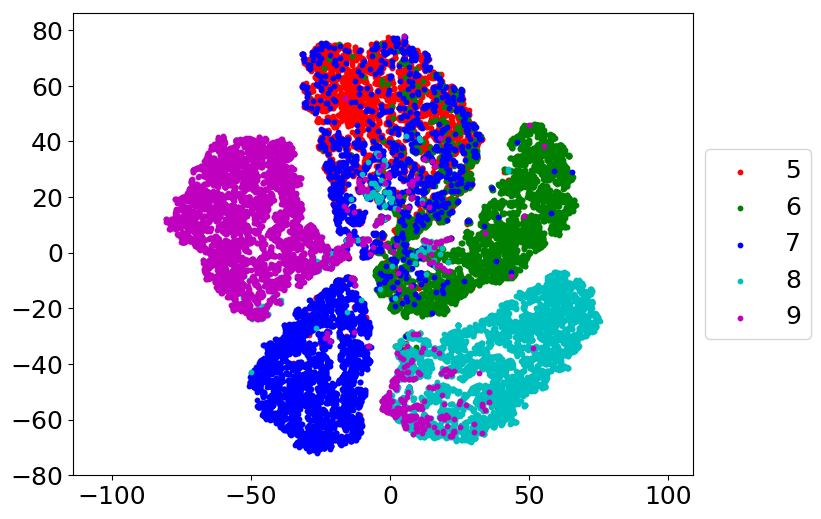} \\
    (j) epoch 80 & (k) epoch 90 & (l) epoch 100\\
   \end{tabular}
    \caption{Representation visualization of unlabelled data (i.e.,  dog, frog, horse, ship, truck) by our deep transfer clustering model. `init' means the feature obtained with the feature extractor trained on the labelled data (i.e., airplane, automobile, bird, cat, deer). Data points are colored according to their ground-truth labels.}
    \label{fig:tsne}
\end{figure*}

We visualize the learned representation of the images in the unlabelled 5-class (i.e.,  dog, frog, horse, ship, truck) subset of CIFAR10 by projecting them to 2D space with t-SNE in~\cref{fig:tsne}. They are with class indices of 5-9 in CIFAR10. 
In~\cref{fig:tsne} (a), we show the representation of unlabelled data obtained by the feature extractor pretrained on labelled data. We can see that, the novel classes can not be properly distinguished, and there are no clear boundaries among different novel classes. In~\cref{fig:tsne} (b)-(l), we show the evolving of the representation at different check points after learning with our model. We can clearly see that the representation becomes more and more discriminative for different classes after learning with our model, clearly demonstrating that our approach can effectively discover novel categories (see~\cref{fig:tsne} (l) vs~\cref{fig:tsne} (a)).

\clearpage

\section{Evolving of soft clustering assignment}
\begin{figure*}[h]
   \centering
   \tabcolsep=0.02cm
   	   \renewcommand{\arraystretch}{0.25}
   \begin{tabular}{
         >{\centering\arraybackslash} m{0.195\textwidth}
         >{\centering\arraybackslash} m{0.20\textwidth}
         >{\centering\arraybackslash} m{0.20\textwidth}
         >{\centering\arraybackslash} m{0.20\textwidth}
         >{\centering\arraybackslash} m{0.20\textwidth}}
    \includegraphics[width=1\linewidth]{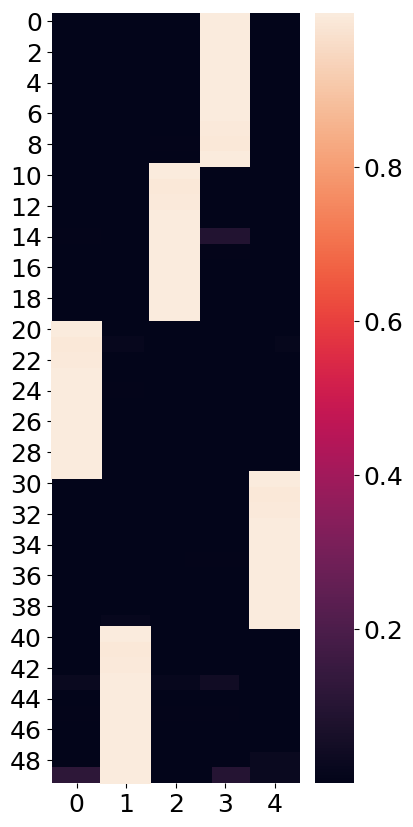} &
    \includegraphics[width=1\linewidth]{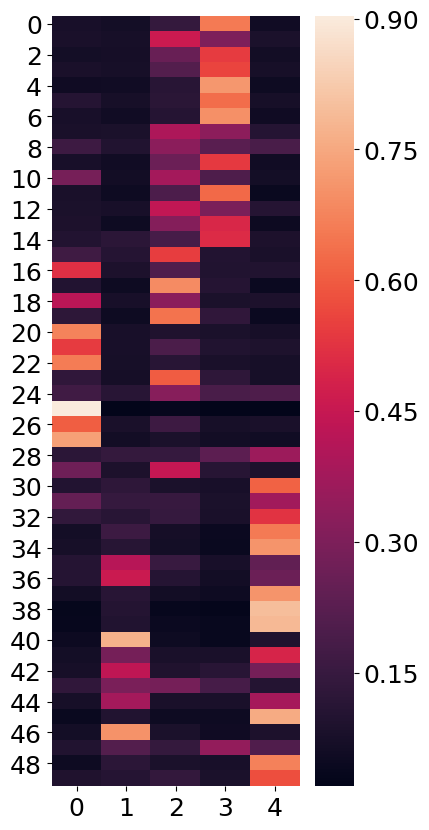} &
    \includegraphics[width=1\linewidth]{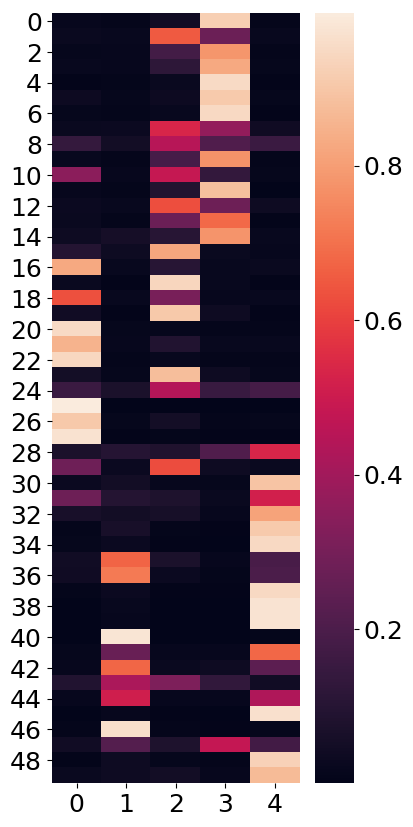} &
    \includegraphics[width=1\linewidth]{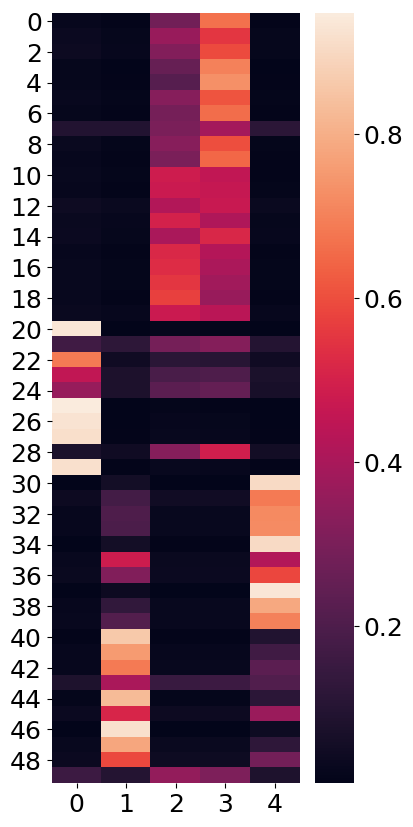} &
    \includegraphics[width=1\linewidth]{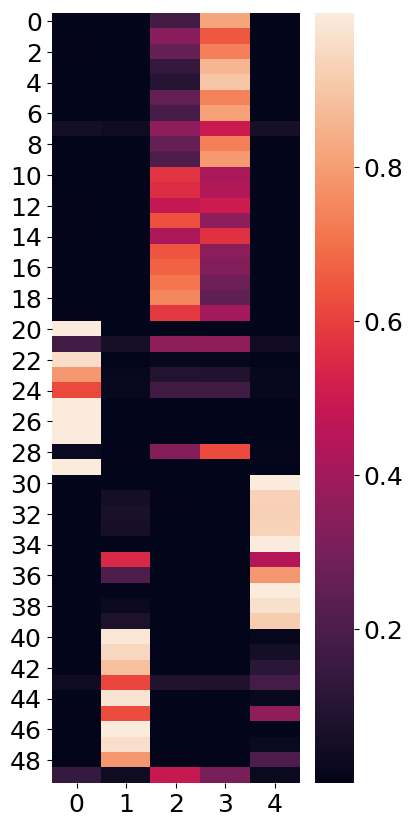} \\
    (a) GT & (b) init: $P$ & (c) init: $Q$ & (d) epoch 20: $P$ & (e) epoch 20: $Q$\\
                                                                          &
    \includegraphics[width=1\linewidth]{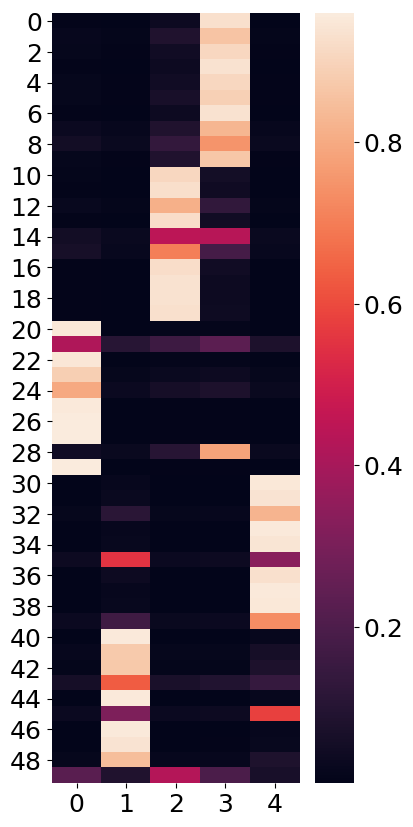} &
    \includegraphics[width=1\linewidth]{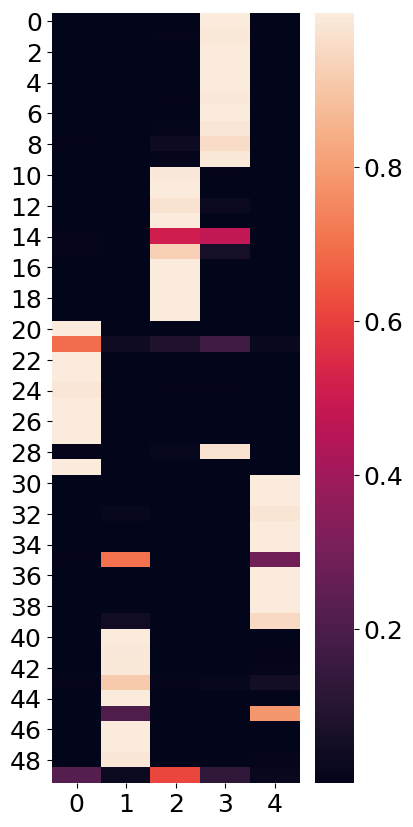} &
    \includegraphics[width=1\linewidth]{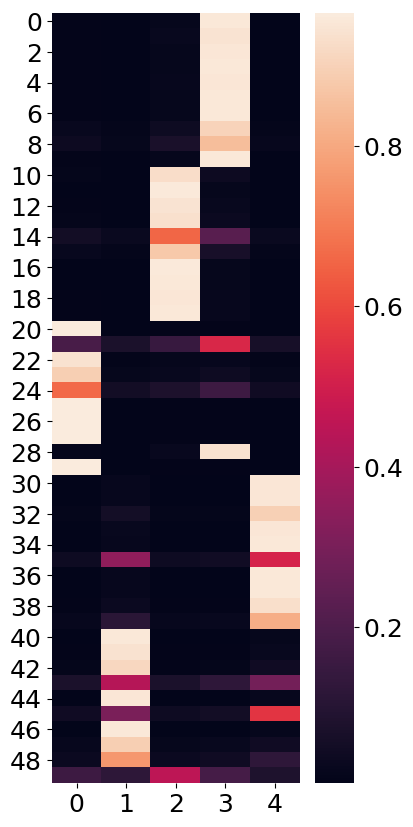} &
    \includegraphics[width=1\linewidth]{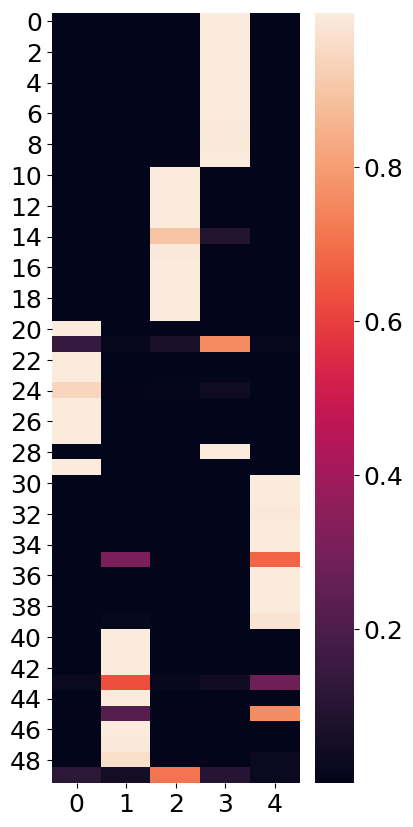} \\
     & (f) epoch 60: $P$ & (g) epoch 60: $Q$ & (h) epoch 100: $P$ & (i) epoch 100: $Q$\\
   \end{tabular}
    \caption{Evolving of soft clustering assignment on unlabelled data. The horizontal axis represents probabilities (i.e., each row represents the probabilities to assign the instance to different clusters), while the vertical axis represents the instance identities. $P$ stands for prediction (soft clustering assignment), and $Q$ stands for constructed target distribution (as described in section 3.1.1 of our manuscript).}
    \label{fig:heatmap}
\end{figure*}

As discussed in section 3.1.1 of our manuscript, the training of our model is driven by the self-evolving soft clustering assignment. By constructing the target distribution $Q$ with the prediction $P$ (soft clustering assignment), our model can gradually learn to discover novel categories. We show how the soft clustering assignment evolves on the unlabelled subset of CIFAR10 in~\cref{fig:heatmap}. Instances with ID 0-9, 10-19, 20-29, 30-39, and 40-49 are images of dog, frog, horse, ship, and truck, respectively. If the images are clustered perfectly, each row in the soft assignment figure will form a one-hot vector, and the predictions will form a vertical bar for each of the 10-instance clusters (see~\cref{fig:heatmap} (a) for the ground truth). As can be seen in~\cref{fig:heatmap} (b), the predictions, which are obtained using the pretrained feature extractor, are rather noisy at the beginning. For example, the model seems unable to distinguish the two classes of instances 0-20 in~\cref{fig:heatmap} (b, d), while after training, our model can perfectly cluster them into two distinct clusters (see~\cref{fig:heatmap}(h)). Similarly, images 40-49 are mostly wrongly considered to be in the same cluster with images 30-39 with high confidence. After training, our model can correctly cluster them. Meanwhile, we can see that the constructed $Q$ is more `confident' and discriminative than $P$, indicating that $Q$ can serve as a proper learning target.

\clearpage
\section{t-SNE visualization with images}

\begin{figure*}[h]
   \centering
   \tabcolsep=0.02cm
   
       \renewcommand{\arraystretch}{0.25}
   \begin{tabular}{
         >{\centering\arraybackslash} m{0.55\textwidth}}
    \includegraphics[width=1\linewidth]{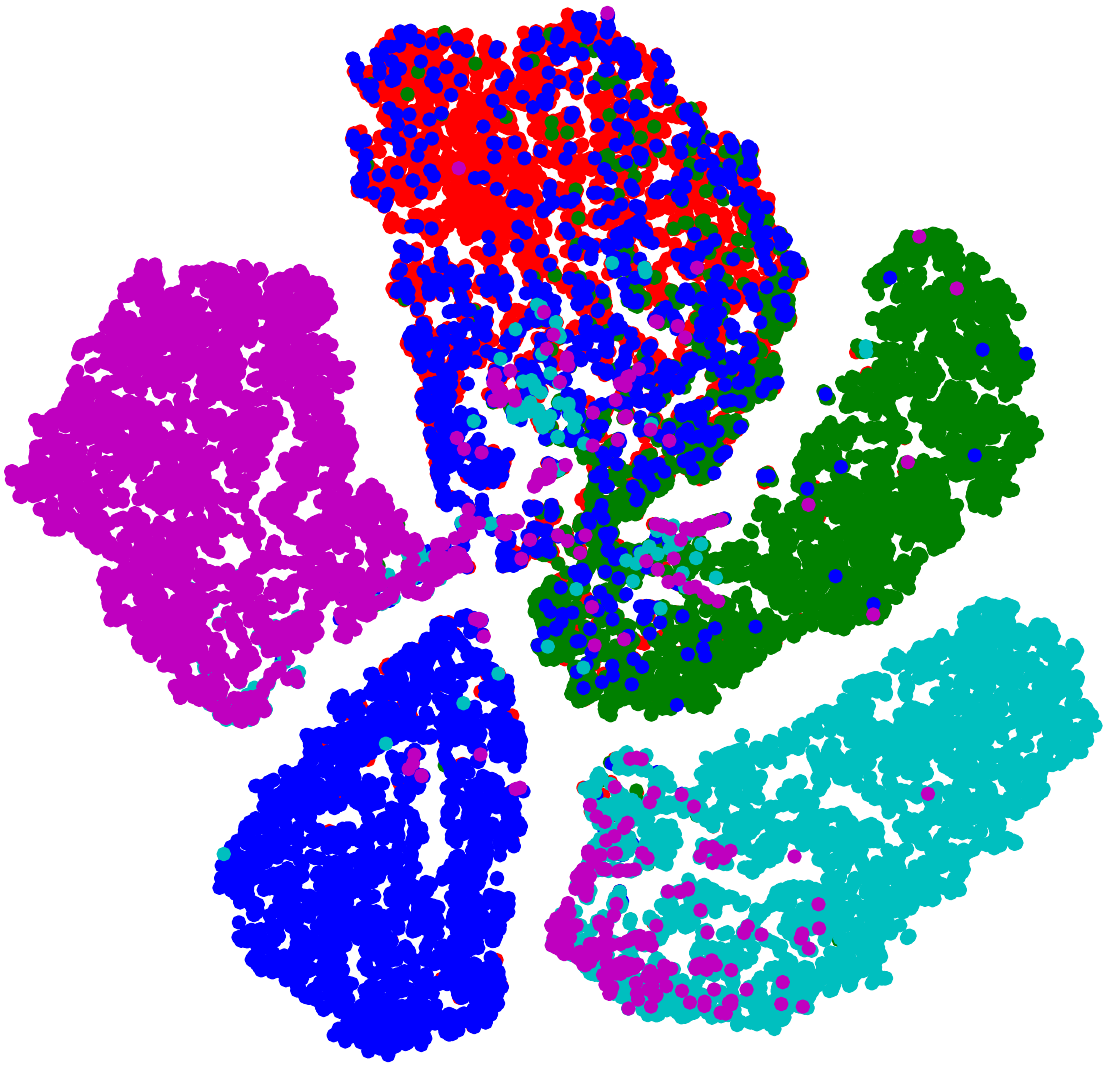}\\
    \includegraphics[width=1\linewidth]{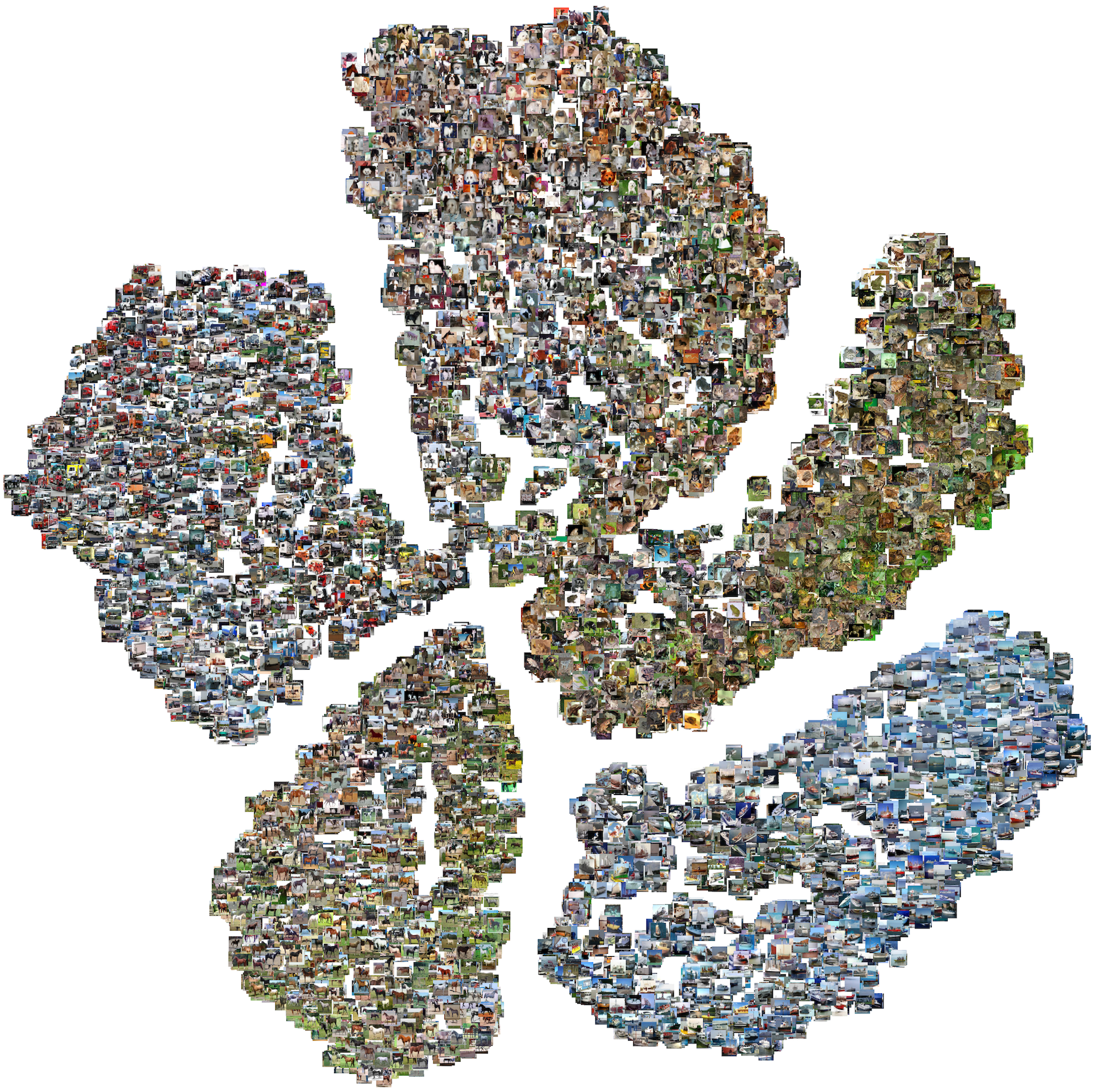}\\
   \end{tabular}
    \caption{Representation visualization of unlabelled subset from CIFAR-10 (i.e.,  dog, frog, horse, ship, truck) by our deep transfer clustering model. Upper: embedding projection colored with GT labels; Lower: the features are associated with their corresponding images.}
    \label{fig:tsne}
\end{figure*}

\clearpage
\section{Detailed category number estimation results on OmniGlot}

KCL~\cite{Hsu18_L2C} amd MCL~\cite{Hsu19_MCL} assume the number of categories to be a large value (i.e., 100 ) instead of estimating the number of categories explicitly.
After running their clustering method, they finally estimate the number of categories by identifying the clusters with a number of assigned instances larger than a certain threshold.
By contrast, we choose to estimate the number of categories before running the transfer clustering algorithm using algorithm 2 in our main paper (with $K_\text{max} = 100$ for all our experiments) and only then apply algorithm 1 to find the clusters.
The results of our estimator for OmniGlot is shown in~\cref{tab:omniglot_k_est}, where we also compared with the results of MCL and KCL as reported in~\cite{Hsu19_MCL}.
Our method achieves better accuracy than the others methods (4.60 vs. 5.10 highest), which validates the effectiveness of our approach.

\begin{table}[ht]
\footnotesize
\centering
\caption{Category number estimation on OmniGlot.}\label{tab:omniglot_k_est}
\begin{tabular}{l|ccccc}
\hline
Alphabet      & GT  & KCL~\cite{Hsu18_L2C} & MCL~\cite{Hsu19_MCL}  & Ours \\
\hline
Angelic & 20 &  26 & 22 & 23\\
Atemayar Q. & 26 & 34 & 26 & 25\\
Atlantean & 26 & 41 & 25 & 34\\
Aurek\_Besh & 26 & 28 & 22 & 34\\
Avesta & 26 & 32 & 23 & 31\\
Ge\_ez & 26 & 32 & 25 & 31\\
Glagolitic & 45 & 45 & 36 & 46\\
Gurmukhi & 45 & 43 & 31 & 34\\
Kannada & 41 & 44 & 30 & 40\\
Keble & 26 & 28 & 23 & 25\\
Malayalam & 47 & 47 & 35 & 42\\
Manipuri & 40 & 41 & 33 & 39\\
Mongolian & 30 & 36 & 29 & 33\\
Old Church S. & 45 & 45 & 38 & 51\\
Oriya & 46 & 49 & 32 & 33\\
Sylheti & 28 & 50 & 30 & 22\\
Syriac\_Serto & 23 & 38 & 24 & 26\\
Tengwar & 25 & 41 & 26 & 28\\
Tibetan & 42 & 42 & 34 & 43\\
ULOG & 26 & 40 & 27 & 33\\
\hline
Avg$_{error}$ & - & 6.35 & 5.10 & \textbf{4.60} \\
\hline
\end{tabular}
\end{table}

\end{document}